\documentclass{article}
\usepackage{acra, balance, epsfig}
\usepackage{amsmath}
\usepackage{autobreak}
\usepackage{graphicx}
\usepackage{tabularx}

\title{Schmidt or Compressed filtering for Visual-Inertial SLAM?}
\author{Hongkyoon Byun and Jonghyuk Kim\\University of Technology Sydney\\
\{Hongkyoon.Byun@student, jonghyuk.kim@\}.uts.edu.au
\AND
Fernando Vanegas and Felipe Gonzalez \\
Queensland University of Technology\\ \{f.vanegasalvarez, felipe.gonzalez\}@qut.edu.au
} 

\begin{document}

\maketitle

\begin{abstract}
Visual-Inertial SLAM has been studied widely due to the advantage of its lightweight, cost-effectiveness, and rich information compared to other sensors. A multi-state constrained filter (MSCKF) and its Schmidt version have been developed to address the computational cost, which treat keyframes as static nuisance parameters, leading to sub-optimal performance. We propose a new Compressed-MSCKF which can achieves improved accuracy with moderate computational costs. By keeping the information gain with compressed form, it can limits to $\mathcal{O}(L)$ with $L$ being the number of local keyframes. The performance of the proposed system has been evaluated using MATLAB simulator.
\end{abstract}

\section{Introduction} 

The autonomous navigation system for mobile robots has been focused on and studied for the past few decades. As a solution, the Simultaneous Localization and Mapping (SLAM) system has been used with different kinds of sensors, such as cameras, Inertial Measurement Units (IMU), Light Detection and Ranging (Lidar), sonar, and so on. Multi-sensor fusion algorithms have been applied to improve the accuracy of this system \cite{RN146}.

Visual-Inertial Navigation System (VINS) has been investigated due to its advantages in size, cost and robustness. The visual sensor itself can detect and track hundreds of features, and it provides rich information through the images, which can bring the improvement of state estimation accuracy. When the camera system loses the tracking of visual features, the IMU can connect the gap with the higher frequency data \cite{qin2019a}.

To resolve the problem of high computational cost, a balance between computational cost and accuracy of the estimation is required for reliable real-time performance \cite{orb-slam}. Multi-State Constraint Kalman Filter (MSCKF) \cite{msckf} delivers the localization information utilising multiple measurements of visual features without including the position of the 3D feature in the filter state vector, which only brings linear computational complexity in the number of features. In Schmidt-MSCKF \cite{schmidt}, key-frame poses are selectively added to the state vector for the loop closure constraints, but without updating the keyframe states to reduce the computational cost. Compressed-SLAM(CP-SLAM) \cite{cpslam} has been demonstrated in which the active state is actively updated within the filter and inactive states are updated in a less frequent rate by accumulating the correlation information.

In this work, we apply the compressed filtering method to the MSCKF to solve the VINS problem, in which the keyframe poses are partitioned into local and global states and updated at a lower rate. In this way, the sub-optimality related to the Schmidt filtering can be adequately handled while constraining the computational complexity. 

The outline of this paper is as follows. Section 2 provides the related work, and Section 3 reviews Schmidt-MSCKF. Section 4 provides the details of the  Compressed-MSCKF and the keyframe vector is updated at a much lower rate by compressing the active and keyframe state correlation information. Section 5 provides experimental results using the high performance MATLAB simulator to verify the Compressed version, followed by conclusion.

\section{Related Work}

Real-time SLAM system in a resource-limited embedded computing system is still challenging in achieving the reliability and robustness.

Extensive research has been conducted to achieve reliable real-time VINS. ORB-SLAM3 \cite{orb3} added the multi-map data association to their previous work \cite{viorb}, which contains ORB sparse front-end, graph optimization back-end, re-localization, and loop closure \cite{review}. It can perform Maximum-a-Posteriori (MAP) estimation during the IMU initialization. Also, the global Bundle Adjustment (BA) is only performed when the keyframe number is below the threshold to reduce the computational complexity. VINS-Mono \cite{qin2017vins} contributed to the robustness of initialization, re-localization, and reusing the pose graph. Their back-end uses a sliding window and selectively marginalizes the state from IMU and features from the sliding window to reduce the computational cost. A lightweight motion-only visual-inertial optimization also has been implemented for the low computational power devices.

In terms of computing resources, EKF based SLAM shows better efficiency in small-end application compared to the above optimization-based methods \cite{review}. \cite{dual-msckf} applied the MSCKF \cite{msckf} as a first layer taking the computational advantage and BA for a loop closure detection in the second layer. It could relieve the linearisation errors and reduce the drift over the long trajectory but still caused the processing overhead during the BA. Maplab \cite{maplab} utilises ROVIOLI \cite{rovioli} and enhanced the online performance, including the tools such as loop-closure, multi-session map merging and pose-graph relaxation. However, this system only yields the accurate pose estimation within the prior map. \cite{schmidt-ekf} implemented the idea from Schmidt Kalman filter \cite{schmidt_original} which selectively including the feature information into the state vector and considers them as a nuisance parameter. There is no updating process for the keyframe state, but cross-correlation with the active state is still maintained. They further applied the method to the MSCKF as \cite{schmidt}. Even though it enables the linear computational complexity, the estimator performance is still sub-optimal due to the static nature of the keyframes.

\section{Schmidt-MSCKF}
In the Schmidt-MSCKF \cite{schmidt}, the standard MSCKF \cite{msckf} is implemented which has the state vector containing the IMU state ${x}_{I}$ and $N$ cloned past camera poses ${x}_{C}$ as a sliding window, 
\begin{flalign}
&{{x}}_{k} =\left[\begin{array}{ll} 
{{x}}_{I}^{\top} & {{x}}_{C}^{\top} \end{array}\right]^{\top} \\
&{x}_{I}=\left[\begin{array}{lllll}
{ }_{G}^{I_{k}} {\bar{q}}^{\top} & {b}_{\omega_{k}}^{\top} & { }^{G} {v}_{I_{k}}^{\top} & {b}_{a_{k}}^{\top} & { }^{G} {p}_{I_{k}}^{\top}\end{array}\right]^{\top}\\
&{{x}}_{C} =\left[\begin{array}{lllll}
{ }_{G}^{C_{k-1}} {\bar{q}}^{\top} & { }^{G} {{p}}_{C_{k-1}}^{\top} &\cdots & {}_{G}^{C_{k-N}} {\bar{q}}^{\top} &{ }^{G} {{p}}_{C_{k-N}}^{\top} \end{array}\right]^{\top}, 
\end{flalign}
where ${ }_{G}^{I} {\bar{q}}$ is the unit quaternion that represent the rotation from frame \{G\} to frame \{I\}, $b_w$ and $b_a$ are the biases of gyroscope and accelerometer, and ${ }^{G} {v}_{I}$ and ${ }^{G} {p}_{I}$ are respectively IMU velocity and position with respect to ${G}$. Furthermore, Schmidt-MSCKF includes the keyframe poses into the state for loop closure as follows, \begin{equation}
\begin{aligned}
{x}_{k}=\left[\begin{array}{lllll}
{x}_{I}^{\top} & {x}_{C}^{\top} & {x}_{S_{1}}^{\top} & \cdots & {x}_{S_{n}}^{\top}
\end{array}\right]^{\top}=\left[\begin{array}{ll}
{x}_{A}^{\top} & {x}_{S}^{\top}
\end{array}\right]^{\top}, 
\end{aligned}
\end{equation}
where the ${{x}_{S_{i}}}=\left[\begin{array}{ll}
{ }_{G}^{C} \bar{q}^{\top} & { }^{G} {p}_{C_{i}}^{\top}
\end{array}\right]^{\top}$ is a keyframe pose to solve the problem of the drifts accumulating over time.

\subsection{Measurement Model}
The measurement model of Schmidt-MSCKF is also derived from the standard MSCKF. In the framework of MSCKF, the camera observations are processed once the features have enough parallax in the sliding window. The same 3D pose point measurements are used to define a constraint equation, relating all the camera poses which are added when there is measurement. Therefore, the stacked nonlinear camera measurement model can be described as \cite{schmidt}:
\begin{flalign}
{z}_{f}={h}\left({x}_{k},{ }^{G} {p}_{f}\right)+{n}_{f},
\end{flalign}
where $n_f$ is the white Gaussian noise with covariance $R_f$, and ${ }^{G} {p}_{f}$ is the 3D position of the feature. The residual model can be obtained using the above measurement model: 
\begin{equation}
\begin{aligned}
{r}_{f} &={z}_{f}-{h}\left(\hat{{x}}_{k \mid k-1},{ }^{G} \hat{{p}}_{f}\right) \\\end{aligned}
\end{equation}
\begin{equation}
\begin{aligned}
{r}_{f} = {H}_{x} \tilde{{x}}_{k \mid k-1}+{H}_{f}{ }^{G} \tilde{{p}}_{f}+{n}_{f},
\end{aligned}
\end{equation}
where ${H}_{x}$ and ${H}_{f}$ are the Jacobians with respect to the state and the feature position, and   $\tilde{{x}}$ and ${ }^{G}\tilde{{p}}$ are the errors of state and feature position. However, standard EKF update cannot be conducted due to the correlation between $\tilde{{x}}$ and ${ }^{G}\tilde{{p}}$. To resolve this problem, $r_f$ is projected to the left nullspace of $H_f$, which can transform into the residual model independent from the position of the feature as: 
\begin{equation}
\begin{aligned}
&{N}^{\top} {r}_{f}={N}^{\top} {H}_{x} \tilde{{x}}_{k \mid k-1}+{N}^{\top} {H}_{f}^{G} \tilde{{p}}_{f}+{N}^{\top} {n}_{f}
\end{aligned}
\end{equation}
\begin{equation}
\begin{aligned}
{r}_{f}^{\prime}={H}_{x}^{\prime} \tilde{{x}}_{k \mid k-1}+{n}_{f}^{\prime},
\end{aligned}
\end{equation}
where ${n}_{f}^{\prime}$ is white Gaussian noise with covariance ${R}_{f}^{\prime}={N}^{\top} {R}_{f} {N}$, and it can be updated as general EKF.

\subsection{Update}
With the derived measurement model, the residual of Schmidt-MSCKF can be described as:
\begin{flalign}
{r}_{f}^{\prime} \simeq {H}_{A_{k}} \tilde{{x}}_{A_{k \mid k-1}}+{H}_{S_{k}} \tilde{{x}}_{S_{k \mid k-1}}+{n}_{f}^{\prime},
\end{flalign}
where $H_{x}^{\prime} = [ H_{A_{k}} H_{S_{k}} ] $. However, due to the keyframes in the state vector, $x_{S}$, which can grow over time, update process cannot be performed in real time. To solve this problem, Schmidt Kalman Filter \cite{schmidt_original} is implemented to avoid the high computational complexity. The intention of SKF update is to treat the keyframe state as static, which makes $K_{S_{k}} $ as a zero. Therefore, using the general EKF update process, the estimation of state can be described as follows:
\begin{align}
\hat{{x}}_{A_{k \mid k}}&=\hat{{x}}_{A_{k \mid k-1}}+{K}_{A_{k}} \tilde{{z}}_{k}^{\prime} \\
\hat{{x}}_{S_{k \mid k}}&=\hat{{x}}_{S_{k \mid k-1}},
\end{align}
where the covariance matrix can be updated keeping the cross-correlation term for consistency as
\begin{equation}
\footnotesize
\begin{aligned}
\begin{array}{l}
{P}_{k \mid k}={P}_{k \mid k-1}- \\
\\
{\left[\begin{array}{ccc}
{K}_{A_{k}} {S}_{k} {K}_{A_{k}}^{\top} & {K}_{A_{k}} {H}_{k}^{\prime} & {\left[\begin{array}{l}
{P}_{A S_{k \mid k-1}} \\
{P}_{S S_{k \mid k-1}}
\end{array}\right]} \\
{\left[\begin{array}{l}
{P}_{A S_{k \mid k-1}} \\
{P}_{S S_{k \mid k-1}}
\end{array}\right]^{\top} {H}_{k}^{\prime}{ }^{\top} {K}_{A_{k}}^{\top}} & 0
\end{array}\right]}
\end{array}. \\
\end{aligned}
\end{equation}
As a result, the computational complexity can be reduced from $O(n^{2})$ to $O(n)$.  

\section{Compressed-MSCKF}

As shown in the previous section, Schmidt-MSCKF implemented the nullspace technique which enables to define the constraints without including the feature position to reduce the computational complexity. In addition, they selectively store keyframes for the loop closure using the Schmidt filtering techniques. It efficiently controls the unbounded localization error and the computational cost by treating the keyframe as static, leading to the linear growth of computational complexity. However, even though it reduces the computational cost, the loss of information cannot be ignored while treating the keyframe state as a 'nuisance' through the trajectory. Finding the right balance between computing requirements and accuracy is a vital part of the real-time SLAM system. Therefore, we present the Compressed-MSCKF including loop closure in this section.

In a framework of standard Compressed SLAM \cite{cpslam}, the state vector is partitioned into a local and global map state. The main benefit of this system is the accumulation of local information gain while operating within the local boundary. This compressed information only propagates to the global state when the local boundary changes to the new boundary, and it allows to update the global map at a much lower rate. In Compressed SLAM, the keyframe states are further partitioned into:
\begin{flalign}
{x}_{S}=\left[\begin{array}{ll}
{x}_{S_{L}}^{\top} & {x}_{S_{G}}^{\top}
\end{array}\right]^{\top},
\end{flalign}
where $x_{S_L}$ and $x_{S_G}$ are the keyframe states in the local and global boundary, respectively. Therefore, we can describe the state vector and the corresponding covariance of the Compressed-MSCKF based on \cite{schmidt} as follow:
\begin{flalign}
{{x_{k}}}=\left[\begin{array}{cc|c}
{{{x}}_{L}} \\
\hline {{{x}}_{G}}
\end{array}\right] = \left[\begin{array}{cc|c}
\begin{array}{ll}
{x}_{A} \\
{x}_{{S}_{L}}
\end{array} \\
\hline {x}_{{S}_{G}}
\end{array}\right]
\end{flalign}
\begin{flalign}
\resizebox{0.8\hsize}{!}{${P}=\left[\begin{array}{l|l}
{P}_{L L} & {P}_{L G} \\
\hline {P}_{G L} & {P}_{G G}
\end{array}\right] = 
\left[\begin{array}{c|c}
\begin{array}{ll}
{P}_{A A} & {P}_{A S_{L}}\\
{P}_{S_{L} A} & {P}_{S_{L} S_{L}}
\end{array} & {P}_{L G}\\
\hline {P}_{G L} & {P}_{G G}
\end{array}\right]$},
\end{flalign}
where ${x}_{A} = \left[\begin{array}{ll} {x}_{I}^{\top} & {x}_{C}^{\top}\end{array}\right]^{\top}$. In this way, since the keyframe need to be kept updated, compression method can be implemented to effectively reduce the computational cost. It can operate the system in $O(L^2)$ of computational complexity with $L$ being the local map size, which is much less than the total map size.

\subsection{Propagation}

Using a standard EKF, the estimation of IMU state can be propagated with the incoming IMU measurement, linear acceleration ($a_m$) and angular velocity ($\omega_m$), based on the following IMU kinematics \cite{msckf}:
\begin{flalign}
 {x}_{k+1}={f}\left({x}_{k}, {a}_{m_{k}}-{n}_{a_{k}}, \boldsymbol{\omega}_{m_{k}}-{n}_{\omega_{k}}\right),
\end{flalign}
where $n_a$ and $n_ {\omega}$ are the zero mean white Gaussian noise of the IMU measurements. The propagation of the covariance matrix also propagates as:
\begin{flalign}
\footnotesize
\begin{array}{l}
{P}_{{L_{k \mid k-1}}}\\
=\left[\begin{array}{c|c}
{J}_{k-1} {P}_{A A} {J}_{k-1}^{T}& {J}_{k-1} {{P}}_{A S_{L}} \\
\hline\left({J}_{k-1} {P}_{A S_{L}}\right)^{T} & {P}_{S_{L} S_{L}}
\end{array}\right] + \left[\begin{array}{c|c}
{Q}_{k-1} & {0} \\
\hline {0} & {0}
\end{array}\right]
\end{array},
\end{flalign}
where $J$ and $Q$ are the system Jacobian and discrete noise covariance matrices for the local state. It can be seen that the correlation between the active and schmidt-keyframe states can be expressed in a compressed form:
\begin{align}
{P}_{A S_{L}}(k)=\left(\prod_{i=1}^{k} {J}_{k-1}\right) {P}_{A S_{L}}(0),
\end{align}
which can be compressed until the local boundary changes.

\subsection{Update}

\begin{figure*}[th!]\centering
\epsfig{figure=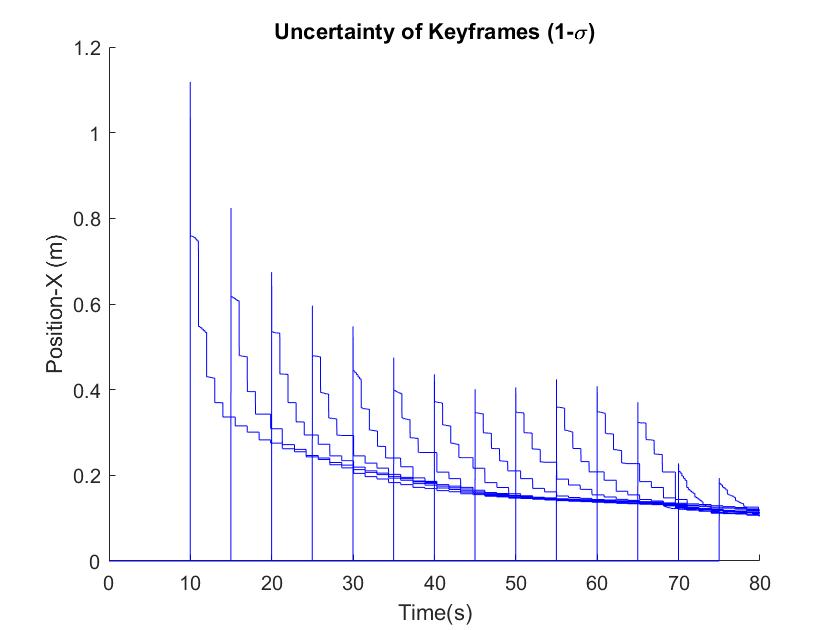,width=0.5\linewidth}
\epsfig{figure=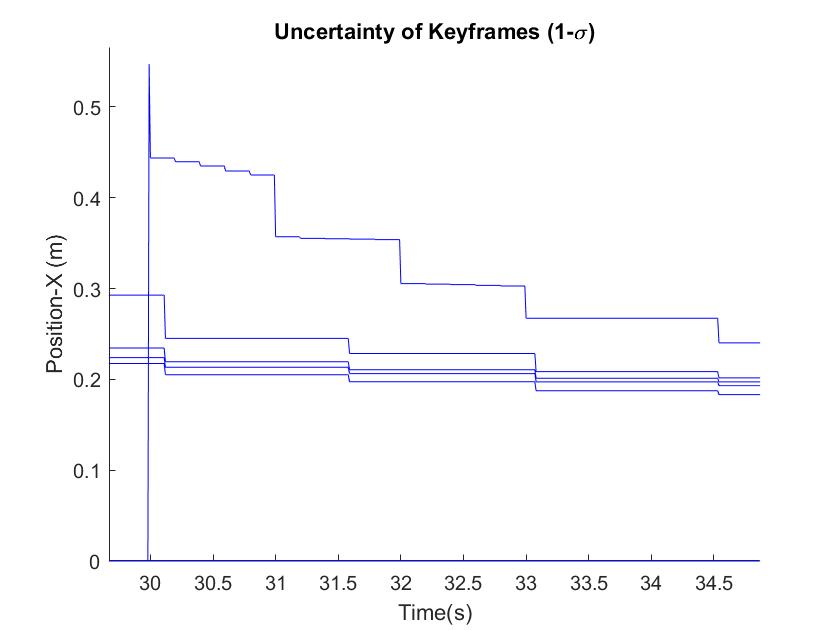,width=0.5\linewidth}
\caption{The evolution of uncertainty of keyframes and the enhanced view of keyframe number 5 showing the compressed updates during the simulation.} \label{fig:1}
\end{figure*}

In our Compressed-MSCKF framework, the observations are first used to update the local state, and the correlation is accumulated. Observation matrix $H$ is sparse and has values only for the local state as ${H^{\prime}}_{x}=\left[\begin{array}{ll}{H}_{L_{k}} &{0}_{G_{k}}\end{array}\right]$ and it can derive the residual measurement as:
\begin{flalign}
{r}_{f}^{\prime} \simeq {H}_{L_{k}} \tilde{{x}}_{L_{k \mid k-1}}+{n}_{f}^{\prime}
\end{flalign}
Using this model, the updating process of state estimate can be derived as follows:
\begin{align}
\hat{{x}}_{L_{k\mid k}} &=\hat{{x}}_{L_{k \mid k-1}}+{K}_{L_{k}}\tilde{{z}}^{\prime}_{k}\\
\hat{{x}}_{G_{k\mid k}} &=\hat{{x}}_{G_{k \mid k-1}}+{K}_{G_{k}}\tilde{{z}}^{\prime}_{k}
\end{align}
The Kalman gain can be computed as
\begin{flalign}
{K_{k}}={P} {H}^{T} {S}^{-1}=\left[\begin{array}{c}
{P}_{L L_{k\mid k-1}} {H}_{L_{k}}^{T} {S_{k}}^{-1} \\
{P}_{G L_{k\mid k-1}} {H}_{L_{k}}^{T} {S_{k}}^{-1}
\end{array}\right]=\left[\begin{array}{l}
{K}_{L_{k}} \nonumber \\
{K}_{G_{k}}
\end{array}\right],
\end{flalign}
where ${S_{k}}={H^{\prime}_{k} P_{k \mid k-1} H^{\prime}_{k}}^{T}+{R^{\prime}}={H}_{L_{k}} {P}_{L L_{k \mid k-1}} {H}_{L_{k}}^{T}+{R^{\prime}}$, and therefore the form of updated covariance matrix is represented as:
\begin{align}
\resizebox{1.\hsize}{!}{$\begin{array}{l}
{P}_{{k \mid k}}={P}_{{k \mid k-1}}-{K}_{{k}} {S_{k}} {K}_{{k}}^{T}\\
\\
={P}_{{k \mid k-1}}-\left[\begin{array}{c}
{P}_{L L_{k \mid k-1}} {H}_{L_{k}}^{T} {S_{k}}^{-1} \\
{P}_{G L_{k \mid k-1}} {H}_{L_{k}}^{T} {S_{k}}^{-1}
\end{array}\right] {S}\left[\begin{array}{c}
{P}_{L L_{k \mid k-1}} {H}_{L_{k}}^{T} {S_{k}}^{-1} \\
{P}_{G L_{k \mid k-1}} {H}_{L_{k}}^{T} {S_{k}}^{-1}
\end{array}\right]^{T}\\
\\
={P}_{{k \mid k-1}}-\\
\left[\begin{array}{cc}
{P}_{L L_{k \mid k-1}}\left({H}_{L_{k}}^{T} {S_{k}}^{-1} {H}_{L_{k}}\right) {P}_{L L_{k \mid k-1}} & \left.(\underbrace{{P}_{L L_{k \mid k-1}}\left({H}_{L_{k}}^{T} {S_{k}}^{-1} {H}_{L_{k}}\right.}_{\phi})\right.) {P}_{L G, k \mid k-1} \\
\left(\left({P}_{L L_{k \mid k-1}}\left({H}_{L_{k}}^{T} {S_{k}}^{-1} {H}_{L_{k}}\right) {P}_{L G_{k \mid k-1}}\right)^{T}\right. & {P}_{G L_{k \mid k-1}}(\underbrace{{H}_{L_{k}}^{T} {S_{k}}^{-1} {H}_{L_{k}}}_{\Psi}) {P}_{L G, k \mid k-1}
\end{array}\right].
\end{array}$}
\end{align}

During this process, the terms of ${K}_{G_{k}}{S}_{k}{K}_{G_{k}}^{\top}$ shows the largest computational cost in standard system and it can be computed in compressed form to effectively handle the computational complexity. Correlation and global state terms can be shortened using ${\phi}$,${\Psi}$, and ${\Phi}$ as follows:   
\begin{flalign}
{P}_{L G_{k \mid k}}={P}_{L G_{k \mid k-1}}-\phi {P}_{L G_{k \mid k-1}}=\Phi {P}_{L G_{k \mid k-1}}\\
{P}_{G G_{k \mid k}}={P}_{G G_{k \mid k-1}}-\left({P}_{G L_{k \mid k-1}} \Psi {P}_{L G_{k \mid k-1}}\right)\\
\hat{{x}}_{{G_{k \mid k}}}={\hat{x}_{G_{k \mid k-1}}}+\left({P}_{G L_{k \mid k-1}} {H}_{L_{k}}^{T} {S_{k}}^{-1} \tilde{{z}}^{\prime}\right)
\end{flalign}
As a result, accumulated form of the global state, correlation term ${P}_{L G}$ ,and global state vector can be expressed as below: 
\begin{equation}
{{P}_{L G}(k)=\left(\prod \Phi\right) {P}_{L G}(0)=\Phi(k, 0) {P}_{L G}(0)}
\end{equation}
\begin{equation}
\resizebox{.85\hsize}{!}{${P}_{G G}(k)={P}_{G G}(0)-{P}_{G L}(0)\left(\sum \Phi(k, 0)^{T} \Psi \Phi(k, 0)\right) {P}_{L G}(0)$}
\end{equation}
\begin{equation}
\hat{{x}}_{{G}}(k)=\hat{{x}}_{{G}}(0)+{P}_{G L}(0)\left(\sum \Phi(k, 0)^{T} {H}_{L}^{T} {S}^{-1} \tilde{{z}}^{\prime} \right)
\end{equation}
Using this compressed correlation term, the global map and covariance can be recovered at a much lower rate whenever the local map boundary changes.

\section{Preliminary Results}

A high-fidelity MATLAB simulator is used to verify the method. The simulator, called Compressed-Pseudo-SLAM, is for all-source navigation and utilises the landmark-based visual-inertial SLAM together as well as GPS pseudorange and pseudorange rate information. The sensor data of IMU (at $100$Hz), vision (at $30$Hz) and pseudoranges (at $1$Hz) are generated following the simulated trajectory (a racehorse track in this work) using realistic sensor models. 

To demonstrate the compressed-MSCKF filter, the simulator is modified to augment the keyframes of the vehicle poses into the state vector. The frequency of the autmentation is dependent on the camera's field-of-view and speed of the vehicle, and a $5$-second interval is used in this work which evenly convers the trajectory. Figure \ref{fig:1} (left) shows the uncertainty evolution of the keyframes showing total $14$ keyframes registered. The compressed update is applied whenever the vehicle reaches the boundary of the local map, in which case the centre of the local map is re-centred at the current vehicle location. In the simulation, the local map is re-centred approximately $1.5-2$ seconds. Figure \ref{fig:1} (right) shows the enhanced view of the keyframe number $5$, showing the effects of sensor updates as well as the compressed updates. At around $30$ seconds, the fifth keyframe is added to the state vector and continuously updated from the vision information. The correction at $31$ seconds is due to the $1$Hz GPS data, while the correction at around $34.5$ seconds is due to the global update. After $2$ global updates, the keyframe becomes out of the local region and becomes an in-active state, corrected only by the global compressed updates. In the Schmidt-MSCKF, the keyframes are initialised using the corresponding vehicle poses, but without the covariance information. A more in-depth analysis is planned to understand the optimal trade-off between the computational complexity and accuracy.    

\section{Conclusions}
In this work, we applied the compressed filtering framework to MSCKF-based Visual-Inertial SLAM. Existing Schmidt-filter based approach can reduce the computational cost by treating the pose keyframes as static variables and thus sacrifysing the accuracy. We showed that we can still retain the pose keyframes in the SLAM state while limiting the computational complexity to $\mathcal{O}(L)$ with $L$ being the number of local keyframes. Future is on the trade-off analysis between the computational cost and accuracy and the implementation of the method for a real dataset collected from drones.



{}

\balance

\end{document}